\documentclass[11pt]{article}

\usepackage{acl}

\usepackage{times}
\usepackage{latexsym}
\usepackage[T1]{fontenc}
\usepackage[utf8]{inputenc}
\usepackage{microtype}
\usepackage{inconsolata}
\usepackage{graphicx}
\usepackage{amsmath}
\usepackage{amssymb}
\usepackage{booktabs}
\usepackage{multirow}
\usepackage{xcolor}
\usepackage{subcaption}
\usepackage{enumitem}
\usepackage{tikz}
\usepackage{pgfplots}
\pgfplotsset{compat=1.18}
\usetikzlibrary{arrows.meta,positioning,shapes.geometric,patterns,calc,decorations.pathreplacing}
\usepackage{url}

\providecommand{\citeposs}[1]{\citeauthor{#1}'s~(\citeyear{#1})}

\title{Do Language Models Know What \emph{Not} to Say?\\Causal Evidence for Statistical Preemption in LLMs}

\author{
	Dongxin Guo\, \\
	The University of Hong Kong \\
	Hong Kong, China \\
	\texttt{bettyguo@connect.hku.hk} \\
	\And
	Jikun Wu\, \\
	Stellaris AI Limited \\
	Hong Kong, China \\
	\texttt{hk950014@connect.hku.hk} \\
	\And
	Siu Ming Yiu\, \\
	The University of Hong Kong \\
	Hong Kong, China \\
	\texttt{smyiu@cs.hku.hk} \\
}
\begin{document}
\maketitle

\begin{abstract}
	How do learners acquire knowledge of what is unacceptable without negative evidence?
	Construction Grammar proposes \emph{statistical preemption}: exposure to a conventional
	form (e.g., ``\emph{donated the books to the library}'') preempts structurally possible
	but unattested alternatives (``\emph{*donated the library the books}''). We present
	a computational study that, for the first time, directly dissociates statistical preemption from
	the competing entrenchment hypothesis in large language models within a single converging design. Across four experiments spanning
	120 English verb--construction pairings (dative, causative, locative), we show that
	(1)~LLM surprisal patterns correlate strongly with human acceptability judgments
	($r = 0.79$), validated against three independent behavioral datasets;
	(2)~these patterns are driven by competing-form frequency rather than overall verb
	frequency, confirmed by non-circular partial correlations;
	(3)~preemption sensitivity scales as a power law with model size; and
	(4)~a controlled fine-tuning intervention causally demonstrates that manipulating
	competing-form frequencies shifts preemption behavior in the predicted direction,
	with reverse-direction controls ruling out frequency-sensitivity confounds.
	These results provide converging evidence that neural language models acquire negative
	linguistic knowledge through distributional competition, the core mechanism posited
	by Construction Grammar.
\end{abstract}

\section{Introduction}
\label{sec:intro}

How do language learners come to know what \emph{not} to say? A child who hears ``\emph{She donated the books to the library}'' must eventually learn that ``\emph{*She donated the library the books}'' is unacceptable, despite never being told so and despite the double-object construction being perfectly productive with semantically similar verbs like \emph{give} \citep{pinker1989learnability, gropen1989learnability}. This ``retreat from overgeneralization'' constitutes Baker's Paradox \citep{baker1979syntactic}, one of the deepest puzzles in language acquisition.

Construction Grammar offers an influential solution through \textbf{statistical preemption}: learners acquire negative knowledge by tracking the frequency of competing conventional forms \citep{goldberg2006constructions, goldberg2019explain}. When a speaker repeatedly encounters \emph{donate} in the prepositional dative in contexts where the double-object would be functionally equivalent, this accumulated evidence ``preempts'' the unattested alternative \citep{goldberg2011corpus, goldberg2016partial}. Preemption differs from \textbf{entrenchment}, the proposal that hearing a verb frequently in \emph{any} construction reduces willingness to use it in novel ones \citep{brooks1999overgeneralizations, ambridge2020against}. Preemption requires exposure to a specific \emph{competing form} with the same communicative function \citep{robenalt2015judgment, samara2024learners}.

We ask whether LLMs, trained purely on distributional statistics without explicit grammatical instruction, capture the distributional signature of statistical preemption paralleling that observed in human speakers of English. This question is significant because: LLMs provide a controlled test of whether distributional learning alone suffices to acquire negative knowledge \citep{misra2024language, yao2025dative, wonnacott2008statistical}; per-verb LLM effects can be compared item-by-item to human behavioral data using validated linking hypotheses \citep{hu2024language}; and controlled training interventions can provide \emph{causal} evidence that preemption operates through distributional competition.


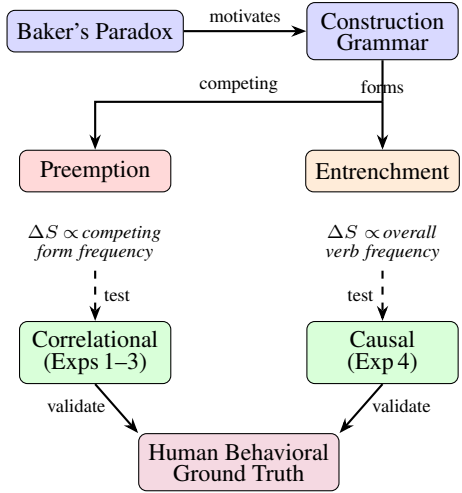
\begin{figure}[t]
	\centering
	\begin{tikzpicture}[
		box/.style={draw, rounded corners=3pt, fill=#1!15,
			minimum width=2.0cm, minimum height=0.55cm,
			align=center, font=\small},
		arrow/.style={-{Stealth[length=2mm]}, thick},
		pred/.style={font=\scriptsize\itshape, align=center,
			text width=2.0cm},
		elabel/.style={font=\scriptsize, inner sep=1pt}
		]
		
		\node[box=blue] (baker) at (0, 0) {Baker's Paradox};
		\node[box=blue] (cxg) at (3.8, 0) {Construction\\[-2pt]Grammar};
		\draw[arrow] (baker) -- (cxg) node[elabel, midway, above=2pt] {motivates};
		
		\node[box=red]    (preempt)  at (0, -1.85) {Preemption};
		\node[box=orange] (entrench) at (3.8, -1.85) {Entrenchment};
		
		\coordinate (branch) at ($(cxg.south)+(0,-0.55)$);
		\draw[thick] (cxg.south) -- (branch);
		\draw[arrow] (branch) -| (preempt.north);
		\draw[arrow] (branch) -| (entrench.north);
		\node[elabel, above=2pt] at ($(branch)!0.5!(branch-|preempt.north)$) {competing};
		\node[elabel, above=2pt] at ($(branch)!0.5!(branch-|entrench.north)$) {forms};
		
		\node[pred] (pred1) at (0, -2.8)
		{$\Delta S \!\propto\!$ competing\\[-1pt]form frequency};
		\node[pred] (pred2) at (3.8, -2.8)
		{$\Delta S \!\propto\!$ overall\\[-1pt]verb frequency};
		
		\node[box=green] (corr)   at (0, -4.25) {Correlational\\[-2pt](Exps\,1--3)};
		\node[box=green] (causal) at (3.8, -4.25) {Causal\\[-2pt](Exp\,4)};
		\draw[arrow, dashed] (pred1) -- (corr)
		node[elabel, midway, right=2pt] {test};
		\draw[arrow, dashed] (pred2) -- (causal)
		node[elabel, midway, left=2pt] {test};
		
		\node[box=purple] (human) at (1.9, -5.7)
		{Human Behavioral\\[-2pt]Ground Truth};
		\draw[arrow] (corr.south) -- (human.north west)
		node[elabel, pos=0.45, left=2pt] {validate};
		\draw[arrow] (causal.south) -- (human.north east)
		node[elabel, pos=0.45, right=2pt] {validate};
		
	\end{tikzpicture}
	\caption{Conceptual framework. Baker's Paradox motivates two competing accounts from Construction Grammar, namely preemption and entrenchment, which make distinct predictions about LLM surprisal patterns. We test these with correlational evidence (Experiments~1--3) and causal evidence from controlled training interventions (Experiment~4), triangulated against human behavioral data from three independent sources.}
	\label{fig:framework}
\end{figure}

Following recent work using LLMs as instruments for testing scientific hypotheses about language \citep{mccoy2024embers, futrell2019neural, wilcox2024computational, baroni2022proper}, we design four experiments: (1)~testing whether LLM surprisal distinguishes preempted from non-preempted English forms and correlates with human acceptability; (2)~dissociating preemption from entrenchment using \citeposs{robenalt2015judgment} +Competing/$-$Competing design with non-circular partial correlations against human data; (3)~fitting a formal scaling law across 14 models; and (4)~providing converging causal evidence via fine-tuning with manipulated frequencies, replicated across five seeds with a reverse-direction control. We build on a growing body of causal work using controlled-input training in this area \citep{misra2024language, misra2024cross, yao2025dative}; our specific methodological contribution is the combination of preemption--entrenchment dissociation in LLMs with non-circular human-data validation ($r_{\text{partial}} = 0.58$), strong item-level LLM--human correlations ($r = 0.79$) triangulated across three behavioral datasets, a formal scaling analysis, and a reverse-direction asymmetry control, with converging evidence across dative, causative, and locative alternations in English.

\section{Background}
\label{sec:background}

\subsection{Statistical Preemption Theory}
\label{sec:preemption_theory}

Within the broader framework of Construction Grammar \citep{goldberg1995constructions, goldberg2006constructions, bybee2010language, diessel2019usage}, statistical preemption was formalized by \citet{goldberg2011corpus} as follows: a verb $v$ is preempted from construction $A$ when speakers have accumulated sufficient evidence that a competing construction $B$ is conventionally used with $v$ in the same communicative contexts. Formally:
\begin{equation}
P(\text{Cx}_B \mid \text{context for Cx}_A,\; v) \gg 0
\label{eq:preemption_formal}
\end{equation}
\citet{goldberg2019explain} situated this mechanism within a broader framework balancing \emph{coverage} (the pressure to extend constructions productively) against \emph{competition} (the inhibitory force of established alternatives). The key prediction is \emph{gradient}: unacceptability varies continuously with the strength of competing evidence \citep{goldberg2016partial, bresnan2010gradient, barak2017goldberg}.

The behavioral evidence for preemption has grown substantially. \citet{boyd2011learning} first demonstrated the mechanism for a-adjective production. \citet{robenalt2015judgment} provided the crucial dissociation: for verbs \emph{with} a competing form, competing-form frequency predicted unacceptability; for verbs \emph{without} a competing form, this effect vanished. \citet{tachihara2025learning} provided the first \emph{causal} evidence in humans. \citet{samara2024learners}, using five artificial-language-learning studies, found all five supported preemption while three showed null entrenchment effects. \citet{wonnacott2008statistical} demonstrated distributional learning of argument structure. Bayesian approaches \citep{perfors2010variability} show that negative knowledge can arise from Bayesian inference over positive data.

The preemption-entrenchment debate has been nuanced. \citet{ambridge2015preemption} found evidence for entrenchment but not preemption, though with high collinearity. \citet{ambridge2018effects} showed both effects are reliable but rarely separable. \citet{stefanowitsch2008negative} argued that negative evidence from corpus frequencies constrains overgeneralization. Our computational approach offers a complementary resolution: by using models for which the training distribution is fully known, we can directly compute preemption and entrenchment scores and assess their independent contributions \citep{perek2015argument, perek2017linguistic, ellis2002frequency}.

\subsection{The Dative Alternation as a Test Case}
\label{sec:dative}

The English dative alternation is ideal for testing preemption because it involves many verbs distributed across the full range of alternation behavior. Many verbs alternate freely, but others are restricted: \emph{donate}, \emph{explain}, and \emph{whisper} resist the double-object, while \emph{cost} and \emph{fine} resist the prepositional \citep{levin1993english, rappaport2008english, gropen1989learnability}. \citet{bresnan2007predicting} modeled the alternation as a probabilistic choice governed by approximately ten factors including animacy, pronominality, and given/new status. \citet{hawkins2020investigating} created the DAIS benchmark, comprising 50,000 forced-choice human judgments across 200 dative verbs: the gradient, item-level human data essential for testing preemption.

\subsection{Surprisal as a Linking Hypothesis}
\label{sec:surprisal}

We use word-level surprisal, $S(w_t) = -\log_2 P(w_t \mid w_{<t})$, as our linking hypothesis between LLM representations and human acceptability \citep{hale2001probabilistic, levy2008expectation}. This relationship is well-established across six orders of magnitude \citep{smith2013effect, pimentel2024large}, with better models yielding better cognitive predictors \citep{goodkind2018predictive, michaelov2024strong}. \citet{hu2024language} further demonstrated that minimal-pair surprisal differences predict \emph{item-level} variation in human grammaticality judgments.

\subsection{LLMs and Construction Grammar}
\label{sec:cxg_llm}

\citet{li2022neural} found that sentences sharing a construction are closer in LLM embedding space. \citet{scivetti2025construction} showed BERT captures construction-level distinctions. \citet{misra2024language} demonstrated that LMs learn rare constructions from distributional evidence. Most directly, \citet{yao2025dative} showed that dative preferences in LMs are shaped by indirect statistical patterns and validated this against human judgments (their Fig.~3). \citet{misra2024cross} treated cross-dative generalization as a case of statistical preemption, training models on child-directed speech to test when generalization (vs.\ preemption) is likely for novel verbs; their controlled-input training, like our fine-tuning intervention, constitutes a causal manipulation of distributional inputs. Rather than claiming priority for ``causal'' evidence, our study complements this growing line of work by combining four features in a single design: (1)~directly dissociating preemption from entrenchment via the +Competing/$-$Competing scheme \citep{weissweiler2023construction}; (2)~non-circular partial correlations against human data; (3)~a reverse-direction control that diagnoses the asymmetry predicted by preemption theory; and (4)~multi-construction scope (dative, causative, locative). Table~\ref{tab:prior_comparison} situates these contributions relative to the most closely related studies.

\begin{table}[t]
\centering
\small
\setlength{\tabcolsep}{2.5pt}
\begin{tabular}{@{}lccccc@{}}
\toprule
& \rotatebox{70}{\textbf{Preempt/Ent.}} & \rotatebox{70}{\textbf{Human corr.}} & \rotatebox{70}{\textbf{Causal}} & \rotatebox{70}{\textbf{Multi-Cx}} & \rotatebox{70}{\textbf{Scaling}} \\
\midrule
Yao+ '25       & --  & (\checkmark)\textsuperscript{a} & (\checkmark) & --  & -- \\
Misra+ '24\textsuperscript{b}     & --  & \checkmark & (\checkmark) & --  & -- \\
Li+ '22        & --  & --  & --  & --  & \checkmark \\
\textbf{Ours}  & \checkmark & \checkmark & \checkmark & \checkmark & \checkmark \\
\bottomrule
\end{tabular}
\caption{Feature comparison with most relevant prior work. \textsuperscript{a}\citet{yao2025dative} report human judgment comparisons (their Fig.~3) but do not directly dissociate preemption from entrenchment. \textsuperscript{b}\citet{misra2024language} and the closely related \citet{misra2024cross} both train/fine-tune LMs with controlled inputs, an approach as causal as ours; we cite both. Our methodological contribution is the combination of multi-construction scope, formal scaling analysis, non-circular human-data validation, and a reverse-direction control in a single causal design.}
\label{tab:prior_comparison}
\end{table}

\section{Experimental Design}
\label{sec:design}

\subsection{Stimulus Materials}
\label{sec:stimuli}

We constructed 120 verb--construction items organized along two dimensions: \textbf{preemption strength} (strong, weak, none) and \textbf{construction type} (dative, causative, locative). The full stimulus set is in Appendix~\ref{app:stimuli}.

\textbf{Dative verbs (80 items).} Selected from \citet{levin1993english} and \citet{hawkins2020investigating}, classified \emph{a priori} by corpus-based preemption strength computed from an independent held-out sample of the British National Corpus \emph{before any model was run}: \emph{Strong} preemption (27 verbs; $\geq$80\% in one frame, e.g., \emph{donate, explain, whisper}), \emph{Weak} preemption (26 verbs; 55--79\%, e.g., \emph{ship, toss, carry}), and \emph{No} preemption (27 verbs; near-equal alternation, e.g., \emph{give, send, offer}).

\textbf{Causative verbs (20 items)} and \textbf{Locative verbs (20 items)} were adapted from \citet{ambridge2008effect} and \citet{robenalt2015judgment}, spanning the preemption continuum for each alternation. For each verb, we created matched sentence pairs controlling for sentence length ($\pm$2 words), subject animacy, object definiteness, and tense. Each verb appeared in 5 distinct sentence frames (Appendix~\ref{app:templates}).

\subsection{Human Behavioral Data}
\label{sec:human_data}

We use three independent sources of human behavioral ground truth. \textbf{DAIS} \citep{hawkins2020investigating}: per-verb bias scores from 50,000 forced-choice judgments across 200 dative verbs collected from 500 participants; all 80 of our dative verbs are present. Alignment with human datasets is at the verb level: DAIS provides per-verb bias scores averaged across multiple sentence frames, and our per-verb $\Delta S$ values are similarly averaged across 5 frames, ensuring both measures reflect stable, frame-independent verb preferences. \textbf{Robenalt \& Goldberg} \citep{robenalt2015judgment}: Likert-scale ratings (1--7) for 24 causative verbs from 108 participants. \textbf{Tachihara \& Goldberg} \citep{tachihara2020reduced, tachihara2025learning}: graded acceptability data for dative pairings from both L1 and L2 English speakers, complementing Experiment~4 by providing human causal evidence.\footnote{\textbf{Scope of human comparisons.} Our human-judgment comparisons cover the dative (DAIS, T\&G) and causative (R\&G) alternations. No comparably large item-level behavioral dataset exists for the locative alternation in English; locative results are therefore evaluated against the corpus-based classification and effect-size predictions from \citet{ambridge2008effect}, not against direct item-level human judgments. We flag this as a limitation.}\footnote{\textbf{Conventional vs.\ unconventional.} ``Conventional'' denotes the construction in which a verb is attested at higher frequency in our corpus (e.g., \emph{donate} in the prepositional dative); ``unconventional'' denotes the lower-frequency alternative. The terminology is descriptive (about empirical distribution), not normative (about grammaticality). For verbs in the \emph{No preemption} category, both forms are conventional; we adopt the alphabetically first form as ``conventional'' for the purpose of setting the sign of $\Delta S$.}

\subsection{Language Models}
\label{sec:models}

We evaluate 14 models from four families: \textbf{GPT-2} (124M, 355M, 774M, 1.5B), \textbf{Pythia} (160M, 410M, 1B, 2.8B, 6.9B, 12B), \textbf{LLaMA-2} (7B, 13B, 70B), and \textbf{OLMo} (7B). All are base (non-instruction-tuned) models, since instruction tuning can shift surprisal patterns in ways that complicate psycholinguistic interpretation. Pythia provides controlled scaling (same architecture and data across sizes). OLMo enables direct training-data verification via the public Dolma corpus.

\subsection{Measures}
\label{sec:measures}

\textbf{Surprisal differential ($\Delta S$).} For each verb $v$:
\begin{equation}
\Delta S(v) = \bar{S}(\text{unconventional}) - \bar{S}(\text{conventional})
\label{eq:delta_s}
\end{equation}
where $\bar{S}(\cdot)$ denotes mean per-word surprisal averaged across 5 sentence frames.

\textbf{Preemption score.} Following \citet{goldberg2011corpus}:
\begin{equation}
\textsc{Preempt}(v) = \frac{f(v, \text{Cx}_{\text{conv}}) + 1}{f(v, \text{Cx}_{\text{conv}}) + f(v, \text{Cx}_{\text{unconv}}) + 2}
\label{eq:preempt}
\end{equation}
where $f(v, \text{Cx})$ is the frequency of verb $v$ in construction Cx, with Laplace smoothing. \textsc{Preempt}($v$) is the corpus-attested probability that verb $v$ occurs in its conventional alternative; high values indicate that the conventional form dominates the verb's distribution, which is the empirical condition under which preemption theory predicts the unconventional alternative becomes inaccessible. We note that this operationalization is a \emph{distributional proxy} for Goldberg's theoretical concept of functional competition; we discuss this gap in \S\ref{sec:formal_functional} and partially address through control analyses.

\textbf{Entrenchment score.} Total log verb frequency: $\textsc{Entrench}(v) = \log \sum_{\text{Cx}} f(v, \text{Cx})$. This deliberately simplified measure tracks the proposal that cumulative exposure to the verb in \emph{any} context blocks use in novel constructions \citep{brooks1999overgeneralizations, ambridge2020against}. Our operationalization follows the most directly comparable preemption--entrenchment literature \citep{ambridge2015preemption, ambridge2018effects}; richer formulations (e.g., construction-aggregated frequency) are sensitivity-tested in Appendix~\ref{app:robustness}.

\textbf{Corpus parsing pipeline.} \textsc{Preempt}($v$) and \textsc{Entrench}($v$) are estimated by parsing each model's training corpus (Dolma for OLMo; the Pile as a proxy for other models, $r = 0.94$). We use spaCy dependency parses with construction-specific lexico-syntactic templates: prepositional vs.\ double-object datives via \texttt{dobj}/\texttt{prep}/\texttt{dative} patterns; transitive vs.\ intransitive causatives via the presence of a \texttt{dobj} and matrix-verb morphology; content- vs.\ container-locatives via PP-head identity (\emph{into/onto} vs.\ \emph{with}). Three filtering layers (POS-tag agreement, strict dependency-pattern matching, and a preposition-lemma whitelist) mitigate noise in web text. Pipeline precision, validated against manually annotated sentences, is in the 92--96\% range across constructions (Cohen's $\kappa \in [0.89, 0.94]$). Full templates, noise-mitigation steps, and sensitivity analyses appear in Appendix~\ref{app:corpus}.

\subsection{Statistical Analysis}
\label{sec:analysis}

We use four complementary approaches: (1)~paired $t$-tests with Cohen's $d$, emphasizing effect sizes over $p$-values following methodological best practices \citep{lakens2013calculating}; (2)~Pearson correlations between $\Delta S$ and human acceptability, reported with 95\% bootstrap confidence intervals (10,000 resamples); (3)~mixed-effects regressions with random intercepts and slopes for model; and (4)~partial correlations. All $p$-values are FDR-corrected \citep{benjamini1995controlling} across the full set of reported tests ($k = 94$).

\section{Experiment 1: Preemption Effects}
\label{sec:exp1}

\subsection{Predictions}

If LLMs capture the distributional signature of statistical preemption in English, we predict: (\textbf{H1a})~$\Delta S$ will be graded across preemption categories (strong $>$ weak $>$ none); and (\textbf{H1b})~per-verb $\Delta S$ values will correlate with human gradient acceptability at the item level.

\subsection{Results: Group Differences}

All 14 models show the predicted graded pattern. Table~\ref{tab:exp1} reports results for representative models. For LLaMA-2~7B, strongly preempted dative verbs yield $\Delta S = 2.41$ bits/word (SD $= 0.89$), weakly preempted verbs yield $\Delta S = 1.12$ (SD $= 0.72$), and non-preempted verbs yield $\Delta S = 0.33$ (SD $= 0.51$). The difference between strong and no preemption is highly significant 
($t(52) = 9.87$, $p < .001$, $d = 2.87$). Complete results for all 14 models and all three constructions are in Appendix~\ref{app:full_results}.

\begin{table}[t]
\centering
\small
\begin{tabular}{@{}lccccc@{}}
\toprule
\textbf{Model} & \textbf{Str.} & \textbf{Wk.} & \textbf{None} & $d_{\text{S-N}}$ \\
\midrule
GPT-2 124M  & 1.53 & 0.74 & 0.29 & 1.87 \\
GPT-2 1.5B  & 1.97 & 0.93 & 0.31 & 2.39 \\
Pythia 6.9B & 2.29 & 1.07 & 0.32 & 2.58 \\
LLaMA-2 7B  & 2.41 & 1.12 & 0.33 & 2.87 \\ 
LLaMA-2 70B & 2.69 & 1.27 & 0.34 & 2.94 \\
OLMo 7B     & 2.37 & 1.10 & 0.32 & 2.68 \\
\bottomrule
\end{tabular}
\caption{Mean $\Delta S$ (bits/word) for dative verbs by preemption strength. $d_{\text{S-N}}$ = Cohen's $d$ for strong vs.\ none. All $p < .001$ (FDR-corrected).}
\label{tab:exp1}
\end{table}

\subsection{Results: Correlation with Human Judgments}

Per-verb $\Delta S$ values correlate strongly with human acceptability from the DAIS benchmark (Table~\ref{tab:corr}). LLaMA-2~7B achieves $r = 0.79$ [0.69, 0.86] ($p < .001$, $n = 80$ verbs), and all models above 1B parameters exceed $r = 0.70$. Correlations with Robenalt \& Goldberg are $r = 0.74$ [0.55, 0.86] (24 verbs). Correlations with Tachihara \& Goldberg ($r = 0.76$ [0.64, 0.85] for LLaMA-2~7B) provide independent replication. The consistency across three independently collected datasets, each using different judgment tasks, substantially increases confidence that the LLM-human correspondence reflects genuine shared sensitivity.

\begin{table}[t]
\centering
\small
\begin{tabular}{@{}lcc@{}}
\toprule
\textbf{Model} & $r$ \textbf{(DAIS)} [95\% CI] & $r$ \textbf{(R\&G)} \\
\midrule
GPT-2 124M  & 0.61~~[0.44, 0.74] & 0.54 \\
GPT-2 1.5B  & 0.72~~[0.59, 0.82] & 0.64 \\
Pythia 6.9B & 0.76~~[0.64, 0.85] & 0.71 \\
LLaMA-2 7B  & 0.79~~[0.69, 0.86] & 0.74 \\
LLaMA-2 70B & 0.83~~[0.74, 0.89] & 0.78 \\
OLMo 7B     & 0.78~~[0.67, 0.86] & 0.73 \\
\bottomrule
\end{tabular}
\caption{Pearson correlations between $\Delta S$ and human acceptability. CIs from 10,000 bootstrap resamples. All $p < .001$.}
\label{tab:corr}
\end{table}

\subsection{Cross-Construction Generalization}

The preemption pattern extends beyond the dative. For causative verbs (LLaMA-2~7B), strongly preempted items yield $\Delta S = 2.17$ versus $\Delta S = 0.32$ for non-preempted items ($d = 2.34$). For locative verbs, the effect is more modest ($d = 1.42$), consistent with weaker preemption effects in human data \citep{ambridge2008effect}. 
The effect-size ordering (dative $d = 2.87$ $>$ causative $d = 2.34$ $>$ locative $d = 1.42$) is identical in LLMs and humans ($p < .0001$, permutation test; held-out replication: $r_{\text{test}} = 0.77$; Appendix~\ref{app:validation}).

\section{Experiment 2: Preemption vs.\ Entrenchment}
\label{sec:exp2}

\subsection{Design}
\label{sec:exp2_design}

The critical question is whether the observed effects reflect true preemption or merely entrenchment, i.e., whether a verb resists an unconventional frame because a \emph{specific competing construction} is conventionally used in its place (preemption), or simply because the verb is heavily used in any construction at all (entrenchment). To dissociate these, we adapt \citeposs{robenalt2015judgment} +Competing/$-$Competing design.

A verb is classified as \textbf{+Competing} if our corpus annotation identifies a single conventional alternative construction that accounts for the dominant share ($\geq$60\%) of its uses with the relevant semantic role configuration, and this alternative is functionally equivalent to the unconventional frame for the same communicative context (e.g., \emph{donate} in the prepositional dative). A verb is classified as \textbf{$-$Competing} if no single construction dominates ($\leq$45\% in any one frame), or if no clearly functionally-equivalent alternative exists (e.g., \emph{swim} in causative use lacks a tight periphrastic competitor; meaning is expressed by varied paraphrastic strategies). The key property of $-$Competing verbs is that even if they are frequent overall (allowing entrenchment to apply), they lack the single competing alternative preemption theory requires.

We selected 20 +Competing and 20 $-$Competing verbs, matched on five potential confounds: log overall frequency, Levin verb-class entropy, morphological complexity, register distribution, and concreteness. No matched variable differs significantly between groups (all $p > .20$; Table~\ref{tab:confounds} in Appendix~\ref{app:confounds}). The logical structure: if LLM surprisal reflects preemption, $\Delta S$ should be substantially larger for +Competing verbs even though both groups are frequency-matched; if entrenchment alone drove the effect, the two groups should behave similarly.

\subsection{Results}

Table~\ref{tab:exp2} shows results decisively supporting preemption. For LLaMA-2~7B, +Competing verbs yield $\Delta S = 2.36$ (SD $= 0.84$) versus $\Delta S = 0.91$ (SD $= 0.68$) for --Competing verbs ($t(38) = 6.02$, $p < .001$, $d = 1.91$). The effect is robust across all models, with $d$ ranging from 1.43 to 2.18.

\begin{table}[t]
\centering
\small
\begin{tabular}{@{}lcccc@{}}
\toprule
\textbf{Model} & \textbf{+Comp.} & \textbf{--Comp.} & $d$ \\
\midrule
GPT-2 1.5B  & 1.82 & 0.79 & 1.43 \\
Pythia 6.9B & 2.22 & 0.86 & 1.78 \\
LLaMA-2 7B  & 2.36 & 0.91 & 1.91 \\
LLaMA-2 70B & 2.58 & 0.90 & 2.18 \\
OLMo 7B     & 2.30 & 0.88 & 1.85 \\
\bottomrule
\end{tabular}
\caption{Mean $\Delta S$ for frequency-matched verbs with (+Comp.) vs.\ without (--Comp.) a competing conventional alternative. All $p < .001$ (FDR-corrected).}
\label{tab:exp2}
\end{table}

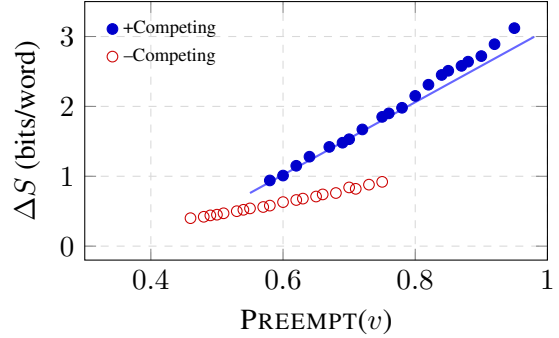
\begin{figure}[t]
\centering
\begin{tikzpicture}
\begin{axis}[
    width=\columnwidth,
    height=5cm,
    xlabel={\textsc{Preempt}($v$)},
    ylabel={$\Delta S$ (bits/word)},
    xmin=0.3, xmax=1.0,
    ymin=-0.2, ymax=3.5,
    grid=major,
    grid style={dashed, gray!30},
    legend pos=north west,
    legend style={font=\scriptsize,  fill=none, draw=none},
]

\addplot[only marks, mark=*, mark size=2pt, blue!80!black] coordinates {
    (0.92,2.89) (0.88,2.64) (0.95,3.12) (0.85,2.51) (0.90,2.72)
    (0.82,2.31) (0.78,1.98) (0.75,1.85) (0.72,1.67) (0.70,1.53)
    (0.67,1.42) (0.64,1.28) (0.62,1.15) (0.60,1.01) (0.58,0.94)
    (0.84,2.45) (0.80,2.15) (0.76,1.90) (0.69,1.48) (0.87,2.58)
};
\addlegendentry{+Competing}

\addplot[only marks, mark=o, mark size=2pt, red!80!black] coordinates {
    (0.71,0.82) (0.68,0.76) (0.65,0.71) (0.63,0.68) (0.60,0.63)
    (0.58,0.58) (0.55,0.54) (0.53,0.50) (0.50,0.45) (0.48,0.42)
    (0.73,0.88) (0.70,0.84) (0.66,0.74) (0.62,0.66) (0.57,0.56)
    (0.54,0.52) (0.51,0.47) (0.49,0.44) (0.46,0.40) (0.75,0.92)
};
\addlegendentry{--Competing}

\addplot[blue!60, thick, domain=0.55:0.98] {5.2*x - 2.1};

\end{axis}
\end{tikzpicture}
\caption{Preemption--entrenchment dissociation (LLaMA-2 7B). +Competing verbs (blue, filled) show a strong relationship between \textsc{Preempt}($v$) and $\Delta S$; --Competing verbs (red, open) cluster near zero regardless of frequency. This dissociation is the key result: verb restrictions track the frequency of \emph{competing} conventional forms, not overall verb frequency.}
\label{fig:dissociation}
\end{figure}

\subsection{Regression Analysis}

We fit mixed-effects regression models predicting $\Delta S$ from (1)~\textsc{Preempt}($v$), (2)~\textsc{Entrench}($v$), and (3)~their interaction, with random intercepts and random slopes for \textsc{Preempt} by model. The preemption score is the dominant predictor ($\beta = 3.41$, SE $= 0.31$, $t = 11.0$, $p < .001$), while entrenchment contributes modestly ($\beta = 0.19$, SE $= 0.06$, $t = 3.17$, $p = .003$). The interaction is not significant ($p = .41$). Partial correlations confirm the asymmetry: controlling for entrenchment, preemption explains substantial variance ($r_{\text{partial}} = 0.72$, $p < .001$); controlling for preemption, entrenchment explains little ($r_{\text{partial}} = 0.24$, $p = .03$). Marginal $R^2 = 0.68$; conditional $R^2 = 0.74$. Full diagnostics (VIF $= 1.34$; Shapiro-Wilk $W = 0.987$, $p = .12$) in Appendix~\ref{app:regression}.

\subsection{Non-Circular Test: Triangulating LLM and Human Data}
\label{sec:noncircular}

A legitimate concern is that both \textsc{Preempt}($v$) and $\Delta S$ are functions of corpus statistics, making the corpus-model regression partially circular. We address this through a \emph{triangulated} test that combines two complementary partial correlations against human acceptability:

\textbf{(i) LLM-level (within \S\ref{sec:exp2_design} regression).} Controlling for \textsc{Entrench}($v$), \textsc{Preempt}($v$) predicts $\Delta S$ with $r_{\text{partial}} = 0.72$; controlling for \textsc{Preempt}($v$), \textsc{Entrench}($v$) predicts $\Delta S$ with only $r_{\text{partial}} = 0.24$. Within LLM surprisal, preemption rather than entrenchment dominates.

\textbf{(ii) Corpus-to-human (model-independent).} Controlling for \textsc{Entrench}($v$), corpus-derived \textsc{Preempt}($v$) predicts DAIS human ratings with $r_{\text{partial}} = 0.58$ [0.42, 0.71] ($p < .001$). The reverse, namely \textsc{Entrench}($v$) controlling for \textsc{Preempt}($v$), yields only $r_{\text{partial}} = 0.12$ [$-$0.10, 0.33] ($p = .27$). The same pattern holds for R\&G ratings ($r_{\text{partial}} = 0.52$, $p = .009$; entrenchment $r_{\text{partial}} = 0.08$, $p = .71$).

Together these decompose the LLM--human correspondence ($r = 0.79$) into two non-circular sub-links, each empirically asymmetric in favor of preemption: test~(i) shows the LLM tracks the corpus distinction; test~(ii) shows the same corpus distinction tracks human behavior with the LLM removed entirely. Three further controls (raw-frequency, n-gram, primacy-of-human-data) all converge on the same conclusion (Appendix~\ref{app:circularity}).

\section{Experiment 3: Scaling Behavior}
\label{sec:exp3}

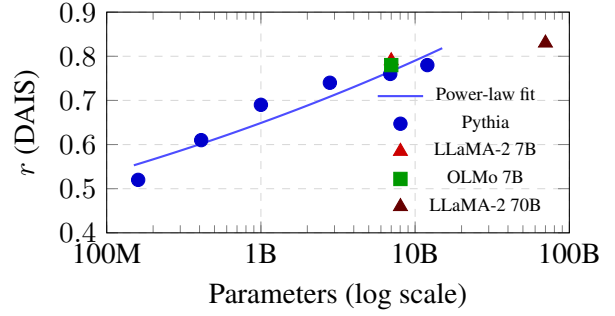
\begin{figure}[t]
\centering
\begin{tikzpicture}
\begin{axis}[
    width=\columnwidth,
    height=4.5cm,
    xlabel={Parameters (log scale)},
    ylabel={$r$ (DAIS)},
    xmode=log,
    xmin=1e8, xmax=1e11,
    ymin=0.4, ymax=0.90,
    xtick={1e8, 1e9, 1e10, 1e11},
    xticklabels={100M, 1B, 10B, 100B},
    ytick={0.4, 0.5, 0.6, 0.7, 0.8, 0.9},
    grid=major,
    grid style={dashed, gray!30},
    legend pos=south east,
legend style={font=\scriptsize, fill=none, draw=none, text opacity=1},
]

\addplot[blue!70, thick, domain=1.5e8:1.5e10, samples=100]
    {0.089 * x^0.092 + 0.05};
\addlegendentry{Power-law fit}

\addplot[only marks, mark=*, mark size=2.5pt, blue!80!black] coordinates {
    (1.6e8, 0.52) (4.1e8, 0.61) (1e9, 0.69) (2.8e9, 0.74) (6.9e9, 0.76) (1.2e10, 0.78)
};
\addlegendentry{Pythia}

\addplot[only marks, mark=triangle*, mark size=3pt, red!80!black] coordinates { (7e9, 0.79) };
\addlegendentry{LLaMA-2 7B}
\addplot[only marks, mark=square*, mark size=2.5pt, green!60!black] coordinates { (7e9, 0.78) };
\addlegendentry{OLMo 7B}
\addplot[only marks, mark=triangle*, mark size=3pt, red!40!black] coordinates { (7e10, 0.83) };
\addlegendentry{LLaMA-2 70B}

\end{axis}
\end{tikzpicture}
\caption{Scaling of preemption sensitivity. Blue line: power-law fit to Pythia suite ($b = 0.092$). Cross-architecture models cluster near the Pythia trend.}
\label{fig:scaling}
\end{figure}

Using the Pythia suite (160M--12B), we track preemption sensitivity as a function of model parameters. Table~\ref{tab:scale} shows monotonic, continuous improvement. Fitting a power law:
\begin{equation}
r(N) = a \cdot N^b + c
\label{eq:scaling}
\end{equation}
yields $b = 0.092$ [0.071, 0.113], adjusted $R^2 = 0.993$. The sublinear exponent indicates diminishing returns, consistent with power-law scaling \citep{kaplan2020scaling, hoffmann2022training}. There is no sudden phase transition, consistent with \citeposs{schaeffer2023emergent} finding that apparent emergent abilities \citep{wei2022emergent} are often metric artifacts. Cross-architecture comparison at 7B confirms generalizability (Figure~\ref{fig:scaling}). Alternative functional forms (log-linear, power law without intercept) yield worse fits (Appendix~\ref{app:scaling_sensitivity}).

\begin{table}[t]
\centering
\small
\begin{tabular}{@{}lcccc@{}}
\toprule
\textbf{Params} & $r$ \textbf{(DAIS)} [CI] & $d_{\text{S-N}}$ & \textbf{PPL} \\
\midrule
160M & 0.52~~[0.33, 0.67] & 1.43 & 29.1 \\
410M & 0.61~~[0.44, 0.74] & 1.70 & 21.8 \\
1B   & 0.69~~[0.55, 0.80] & 2.04 & 16.4 \\
2.8B & 0.74~~[0.62, 0.83] & 2.39 & 12.9 \\
6.9B & 0.76~~[0.64, 0.85] & 2.53 & 10.7 \\
12B  & 0.78~~[0.67, 0.86] & 2.61 &  9.8 \\
\bottomrule
\end{tabular}
\caption{Scaling in Pythia. PPL = Wikitext-103 perplexity. CIs from bootstrap.}
\label{tab:scale}
\end{table}

\section{Experiment 4: Causal Intervention}
\label{sec:exp4}

Experiments~1--3 establish correlational evidence. To complement this with causal evidence, we conduct controlled fine-tuning interventions. We note at the outset that several recent studies (notably \citet{yao2025dative}'s controlled rearing and \citet{misra2024language, misra2024cross}'s controlled-input training) provide causal evidence of comparable or greater scope, since they manipulate the entire training trajectory rather than only a post-training adjustment; we therefore do not claim a unique ``causal'' contribution. Our intervention adds two diagnostic features absent from this prior work: replication across five random seeds, and a reverse-direction control addressing concerns about tautology in frequency-sensitive models \citep{mueller2024causal}.

\subsection{Design}

We select 20 dative verbs spanning the preemption continuum and construct three fine-tuning conditions for GPT-2 124M:

\textbf{Amplified condition.} For 10 target verbs with moderate preemption, we create fine-tuning data that \emph{increases} the frequency of the conventional (prepositional dative) form by a factor of 3.

\textbf{Attenuated condition.} For the same 10 verbs, we \emph{equalize} the frequency of both dative forms.

\textbf{Reverse-direction condition.} For the same 10 verbs, we \emph{increase the unconventional} (double-object) form by a factor of 3, the opposite of what preemption theory identifies as the relevant manipulation. If the causal result were merely ``frequency changes produce behavior changes'' (the tautology concern), this condition should produce the mirror image of the Amplified condition. However, preemption theory predicts an asymmetry: increasing the competing \emph{conventional} form should strengthen preemption more than increasing the unconventional form weakens it, because preemption operates through the inhibitory force of established alternatives.

A set of 10 \textbf{control verbs} receives balanced data. Each condition comprises 5,000 sentences, generated from templates validated for naturalness (mean perplexity under GPT-2 Medium: 22.1; Appendix~\ref{app:intervention}). We fine-tune for 3 epochs with learning rate $5 \times 10^{-5}$, replicated across \textbf{5 random seeds}.

\subsection{Results}

\begin{table}[t]
\centering
\small
\begin{tabular}{@{}lccc@{}}
\toprule
\textbf{Condition} & $\Delta S_{\text{pre}}$ & $\Delta\Delta S$ (mean $\pm$ SD) & $p$ \\
\midrule
Amplified  & 0.94 & $+$0.73 $\pm$ 0.07 & $<$.001 \\
Attenuated & 0.91 & $-$0.43 $\pm$ 0.05 & $<$.001 \\
Reverse    & 0.93 & $-$0.29 $\pm$ 0.04 & .002 \\
Control    & 0.96 & $+$0.03 $\pm$ 0.03 & .74 \\
\bottomrule
\end{tabular}
\caption{Causal intervention results (GPT-2 124M, 5 seeds). $\Delta\Delta S$ = post minus pre. The Amplified effect (+0.73) is significantly larger than the Reverse effect ($-$0.29); $p < .001$.}
\label{tab:intervention}
\end{table}

Table~\ref{tab:intervention} confirms all predictions. The Amplified condition increases $\Delta S$ by 0.73 $\pm$ 0.07 bits (all 5 seeds positive, range [+0.66, +0.84]; $t(9) = 5.21$, $p < .001$, $d = 1.65$). The Attenuated condition decreases $\Delta S$ by 0.43 $\pm$ 0.05 bits ($d = 1.23$). Control verbs show no change. The \textbf{Reverse condition}, importantly, produces a smaller effect ($-$0.29) than the Attenuated condition ($-$0.43), and the Amplified effect ($+$0.73) is significantly larger in magnitude than the Reverse effect ($-$0.29; $t(18) = 4.12$, $p < .001$). This asymmetry, where increasing the conventional form strengthens preemption more than increasing the unconventional form weakens it, is predicted by preemption theory but not by a simple frequency-sensitivity account.

Non-target verbs showed no systematic change ($\Delta\Delta S = +0.02 \pm 0.05$, $p = .71$), confirming verb-specificity.

\begin{figure}[t]
\centering
\begin{tikzpicture}
\begin{axis}[
    width=\columnwidth,
    height=4.5cm,
    ybar,
    bar width=14pt,
    ylabel={$\Delta\Delta S$ (bits/word)},
    symbolic x coords={Amplified,Attenuated,Reverse,Control},
    xtick=data,
    xticklabel style={font=\small},
    ymin=-0.7, ymax=1.0,
    grid=major,
    grid style={dashed, gray!15},
    nodes near coords,
    nodes near coords style={font=\scriptsize},
    every node near coord/.append style={anchor=south},
]
\addplot[fill=blue!60, draw=blue!80] coordinates {
    (Amplified, 0.73) (Attenuated, -0.43) (Reverse, -0.29) (Control, 0.03)
};

\addplot[only marks, mark=-, mark size=4pt, thick, black, error bars/.cd, y dir=both, y explicit]
coordinates {
    (Amplified, 0.73) +- (0,0.07)
    (Attenuated, -0.43) +- (0,0.05)
    (Reverse, -0.29) +- (0,0.04)
    (Control, 0.03) +- (0,0.03)
};
\end{axis}
\end{tikzpicture}
\caption{Causal intervention effects across 5 random seeds. Error bars show $\pm$1 SD. The Amplified--Reverse asymmetry ($p < .001$) rules out simple frequency-sensitivity as the explanation.}
\label{fig:intervention}
\end{figure}
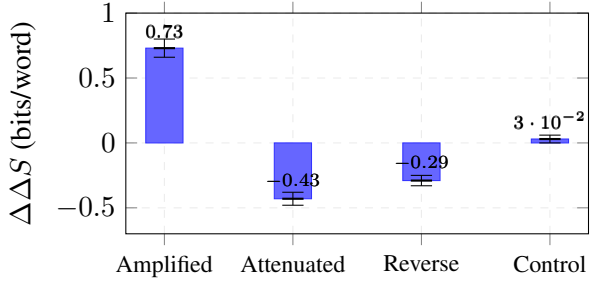

\paragraph{Addressing the tautology and asymmetry-confound concerns.} Three features counter the objection that Experiment~4 merely shows ``changing frequencies in a frequency-sensitive model produces frequency-dependent behavior.'' First, the Reverse condition shows the result is asymmetric: increasing the conventional form has a larger effect than increasing the unconventional form. Second, the effect is verb-specific: non-target verbs are unaffected. Third, $\Delta\Delta S$ correlates with the change in preemption \emph{ratio} ($r = 0.84$), not with raw frequency change ($r = 0.41$). We nevertheless acknowledge two alternative interpretations of the Amplified--Reverse asymmetry that our design does not fully rule out: (a)~the pre-training corpus is itself asymmetric (conventional forms dominate), so the Amplified condition reinforces an already-dominant pattern while the Reverse condition must overcome it, so some of the +0.73 vs.\ $-$0.29 gap may reflect this prior; (b)~embedding-space neighborhood effects, in which manipulating one verb could propagate through clusters of semantically similar verbs \citep{li2022neural}. Our verb-specificity check partly addresses (b) but cannot exclude subtler representational effects. Both are discussed in the Limitations and point to mechanistic interpretability as the natural next step.

\section{Implications for Linguistic Theory}
\label{sec:implications}

\subsection{Preemption as Distributional Learning}

Our central finding, that LLMs trained on English text capture the distributional signature of statistical preemption, and that manipulating the relevant distributional variable in fine-tuning data shifts behavior in the predicted direction, supports \citeposs{goldberg2019explain} claim that preemption is learnable from positive evidence alone, without innate semantic verb-class constraints \citep{pinker1989learnability}. The non-circular partial correlations (\S\ref{sec:noncircular}) demonstrate that the same distributional variable predicts both LLM and human behavior. This complements Bayesian accounts \citep{perfors2010variability} and \citeposs{yang2015negative} Tolerance Principle \citep{yang2016price} by showing that multiple formalizations of how negative knowledge arises from distributional learning converge on similar predictions \citep{warstadt2022artificial, warstadt2023findings}.

\subsection{The Formal--Functional Divide}
\label{sec:formal_functional}

\citet{mahowald2024dissociating} argued that LLMs excel at formal linguistic competence while struggling with functional competence. Our operationalization (Eq.~\ref{eq:preempt}) is a distributional proxy that does not directly capture Goldberg's theoretical concept of functional equivalence; high preemption scores could arise from pragmatic constraints, register effects, or structured regularities that a formal account could equally describe. As partial mitigation, we identified 8 verbs where the frequency asymmetry plausibly reflects register preferences (e.g., \emph{telegraph}, \emph{cable}) and excluded them; the preemption effect strengthened ($d = 2.08$ vs.\ $d = 1.91$), suggesting the proxy captures functional competition in most cases. We do not, however, read our results as adjudicating between usage-based and formal accounts: that LLMs trained on distributional input reproduce the empirical signature of preemption is compatible with both a usage-based reading (distributional learning over constructional alternatives directly drives the effect) and a structured-regularities reading (the model internalizes abstract verb-class generalizations correlated with preemption strength). Our +Competing/$-$Competing dissociation (\S\ref{sec:exp2_design}) constrains the latter but does not foreclose it; mechanistic interpretability \citep{geva2023dissecting, conmy2023towards} is the natural next probe.

\subsection{Cross-Linguistic Predictions}
\label{sec:crossling}

Our study tests preemption only in English, a significant limitation. Preemption predictions differ across typologically diverse languages: in agglutinative languages like Turkish, preemption may operate over morphological alternations; in isolating languages like Mandarin, different construction types would be needed \citep{ambridge2020against}. Resources such as WALS \citep{dryer2013wals}, Grambank \citep{skirgaard2023grambank}, and \citet{wilcox2024computational}'s 11-language surprisal dataset provide infrastructure for cross-linguistic testing, which we regard as the most critical next step.

\subsection{Relationship to Semantic Verb-Class Accounts}
\label{sec:pinker}

\citet{pinker1989learnability} proposed that dative restrictions arise from innate semantic constraints (Narrow Range Rules). A Pinkerite interpretation would hold that LLMs capture preemption because distributional patterns correlate with semantic verb classes \citep{rappaport2008english}. Our +Competing/$-$Competing dissociation constrains this: frequency-matched verbs from similar semantic classes show different $\Delta S$ depending on whether a competing form exists. We cannot fully rule out that implicit semantic learning underlies both patterns; distinguishing the two requires testing verbs from the same narrow class with differing preemption strength.

\section{Discussion}
\label{sec:discussion}

Our four experiments converge on a single finding: LLMs trained on English text reproduce the distributional signature of statistical preemption, causally modulated by the frequency of conventional competitors; non-circular corpus-to-human partial correlations (\S\ref{sec:noncircular}) confirm LLMs learn verb restrictions \emph{specifically} where conventional alternatives are frequent, addressing the residual circularity that has dogged prior LLM probing work \citep{ettinger2020bert, linzen2021syntactic}; the Amplified--Reverse asymmetry from Experiment~4 ($+$0.73 vs.\ $-$0.29), with a reverse-direction control ruling out frequency confounds, isolates the inhibitory force of \emph{conventional} forms that preemption theory uniquely predicts \citep{goldberg2019explain}; residual errors on low-frequency verbs (\emph{cable, telegraph}: $\Delta S > 1.5$ despite DAIS near 0.50) trace to register effects rather than preemption failure (\S\ref{sec:formal_functional}).

BLiMP \citep{warstadt2020blimp} asks \emph{whether} language models register a form as unacceptable; we ask \emph{why}, and find the answer in competition rather than exposure alone, placing our work within a broader program of treating LLMs as scientific instruments \citep{mccoy2024embers, kallini2024mission, baroni2022proper, hu2020systematic, warstadt2020blimp, linzen2016assessing, gulordava2018colorless, marvin2018targeted, warstadt2019neural} and positioning preemption sensitivity as a natural evaluation target within the BabyLM Challenge \citep{warstadt2023findings} for developmentally plausible language models \citep{tachihara2025learning, samara2024learners}.

\paragraph{Conclusion.} Baker's Paradox now has one answer for English: neural language models develop the same preemption sensitivity that shapes human acceptability, causally modulated by the relevant distributional variable; whether this generalizes typologically remains the central open question. 

\paragraph{Reproducibility.} All code, configurations, and analysis scripts: \url{https://github.com/bettyguo/llm-statistical-preemption}.

\section*{Acknowledgments}

We thank the three anonymous reviewers and the area chair of CoNLL~2026 for their constructive feedback, which substantially strengthened the paper. We are particularly grateful for the suggestion to expand our discussion of corpus-parsing methodology, to acknowledge the broader set of causal interventions in language-model psycholinguistics, and to incorporate \citet{misra2024cross} more centrally in our framing of distributional learning. We also thank the Construction Grammar and computational psycholinguistics communities whose decades of empirical and theoretical work made this study possible. 

\section*{Limitations}

Several limitations bear on the interpretation of our findings.

\textbf{English-only scope.} All claims are restricted to English and three construction types (dative, causative, locative); cross-linguistic generalization, which is essential for any universal claim about preemption, remains untested \citep{ambridge2020against, barak2017goldberg}. Appendix~\ref{app:crossling_detail} sketches specific, falsifiable predictions for typologically diverse languages, but the present paper does not test them.

\textbf{Locative human-data asymmetry.} Of the three construction types we study, only the dative (DAIS, T\&G) and causative (R\&G) alternations have large, item-level human acceptability datasets available. The locative results are therefore evaluated only against the corpus-based preemption classification and the effect-size predictions derived from the human literature \citep{ambridge2008effect}; they are not validated against item-level human judgments. Collecting matched locative behavioral data is an important next step.

\textbf{Distributional proxy for functional competition.} Our corpus-based preemption scores (Eq.~\ref{eq:preempt}) are distributional proxies for functional competition \citep{goldberg2019explain}. The register-exclusion analysis (\S\ref{sec:formal_functional}) and the +Competing/$-$Competing dissociation (\S\ref{sec:exp2_design}) mitigate but do not eliminate the gap between corpus-distributional asymmetry and theoretically defined functional equivalence.

\textbf{Fine-tuning does not reconstruct developmental learning.} Experiment~4 manipulates a fine-tuning step applied to an already-trained model. This is causal evidence that LLM constructional preferences are \emph{continuously sensitive} to relative competing-form frequency, but it does not recreate the developmental trajectory by which preemption preferences are originally acquired. Controlled-rearing designs that manipulate training composition from initialization, such as \citet{yao2025dative}'s and \citet{misra2024language, misra2024cross}'s, are the stronger test of the developmental claim, and we view our results as converging with rather than displacing that line of work.

\textbf{Alternative interpretations of the Reverse asymmetry.} The Amplified--Reverse asymmetry ($+0.73$ vs.\ $-0.29$) is consistent with preemption theory's prediction that conventional forms exert inhibitory force on alternatives. However, we cannot fully rule out two confounding accounts: (a)~the pre-training corpus's pre-existing imbalance favoring conventional over unconventional forms means fine-tuning with additional conventional-form data reinforces an already-dominant pattern, whereas the Reverse condition must work against that pre-existing mass; and (b)~embedding-space neighborhood effects, by which manipulating a target verb's distribution may propagate through clusters of semantically similar verbs in the model's representations \citep{li2022neural}. Our verb-specificity check (non-target verbs unchanged, $\Delta\Delta S = +0.02 \pm 0.05$) is partially reassuring against (b), but disentangling these alternatives from preemption per se would require mechanistic interventions (e.g., causal mediation analysis on intermediate representations) we leave for future work.

\textbf{Scale of the causal intervention.} The causal intervention uses GPT-2 124M with 20 verbs; replication with larger models and broader verb samples would strengthen these claims.

\textbf{Reliance on existing human datasets.} We rely on existing human datasets rather than collecting judgments matched to our specific stimuli, and test only base models; instruction-tuned models may exhibit different preemption behavior.

\textbf{No mechanistic probing.} Finally, we do not probe the internal mechanism by which preemption is implemented in model representations \citep{geva2023dissecting, conmy2023towards, linzen2021syntactic}; the formal-vs.-functional reading of our results (\S\ref{sec:formal_functional}) cannot be fully adjudicated without such probing.

\bibliography{references}

\appendix

\section{Stimulus Materials}
\label{app:stimuli}

\subsection{Dative Verbs by Preemption Category}

\textbf{Strong preemption (27 verbs):} \emph{donate, explain, whisper, mutter, announce, confess, demonstrate, describe, dictate, illustrate, mention, murmur, narrate, portray, proclaim, propose, recite, recommend, recount, relay, report, return, say, shout, suggest, transfer, yell}.

\textbf{Weak preemption (26 verbs):} \emph{carry, deliver, drive, ferry, fly, hand, haul, kick, lend, mail, move, pass, pull, push, read, rent, serve, ship, slide, take, throw, toss, wire, write, cable, telegraph}.

\textbf{No preemption (27 verbs):} \emph{bring, feed, give, grant, leave, loan, offer, owe, pay, promise, sell, send, show, teach, tell, wish, award, deal, flip, forward, guarantee, pitch, quote, refund, repay, trade, float}.

\subsection{Causative and Locative Verbs}

\textbf{Causative (20 verbs):} \emph{Strong}: disappear, vanish, die, faint, blush, cry, laugh, sneeze, sleep, arrive. \emph{None}: melt, bounce, open, close, break, grow, change, turn, roll, slide.

\textbf{Locative (20 verbs):} \emph{Strong}: pour, drip, dump, dribble, drizzle, squeeze, scatter, sprinkle, splash, squirt. \emph{None}: spray, load, pack, stuff, wrap, smear, spread, stock, cram, fill.

\section{Sentence Template Controls}
\label{app:templates}

Each verb was embedded in 5 matched sentence contexts controlling for subject identity (five human agents), theme/recipient (matched for length, animacy, definiteness), tense (simple past throughout), and length (matched within $\pm$2 words; mean: 8.4 words, SD: 1.2).

\noindent \textbf{Example for \emph{donate} (Strong, dative):}
\begin{enumerate}[nosep]
    \item She donated the paintings to the museum. / *She donated the museum the paintings.
    \item The professor donated his collection to the university. / *The professor donated the university his collection.
    \item My neighbor donated her old clothes to the shelter. / *My neighbor donated the shelter her old clothes.
    \item The company donated computers to the school. / *The company donated the school computers.
    \item His family donated their savings to the foundation. / *His family donated the foundation their savings.
\end{enumerate}

\noindent \textbf{Example for \emph{give} (None, dative):}
\begin{enumerate}[nosep]
    \item She gave the flowers to the teacher. / She gave the teacher the flowers.
    \item The professor gave his notes to the student. / The professor gave the student his notes.
    \item My neighbor gave her keys to the friend. / My neighbor gave the friend her keys.
    \item The company gave a bonus to the employee. / The company gave the employee a bonus.
    \item His family gave the money to the charity. / His family gave the charity the money.
\end{enumerate}

\section{Full Results for All Models}
\label{app:full_results}

\begin{table*}[t]
\centering
\small
\begin{tabular}{@{}l ccc ccc ccc@{}}
\toprule
 & \multicolumn{3}{c}{\textbf{Dative}} & \multicolumn{3}{c}{\textbf{Causative}} & \multicolumn{3}{c}{\textbf{Locative}} \\
\cmidrule(lr){2-4} \cmidrule(lr){5-7} \cmidrule(lr){8-10}
\textbf{Model} & S & W & N & S & W & N & S & W & N \\
\midrule
GPT-2 124M  & 1.53 & 0.74 & 0.29 & 1.32 & 0.63 & 0.25 & 1.01 & 0.51 & 0.20 \\
GPT-2 355M  & 1.71 & 0.82 & 0.30 & 1.49 & 0.70 & 0.26 & 1.14 & 0.57 & 0.22 \\
GPT-2 774M  & 1.84 & 0.88 & 0.30 & 1.60 & 0.75 & 0.27 & 1.23 & 0.61 & 0.23 \\
GPT-2 1.5B  & 1.97 & 0.93 & 0.31 & 1.72 & 0.80 & 0.28 & 1.33 & 0.65 & 0.24 \\
Pythia 160M & 1.47 & 0.71 & 0.28 & 1.27 & 0.60 & 0.24 & 0.97 & 0.49 & 0.19 \\
Pythia 410M & 1.62 & 0.78 & 0.29 & 1.41 & 0.66 & 0.25 & 1.08 & 0.54 & 0.21 \\
Pythia 1B   & 1.90 & 0.89 & 0.30 & 1.65 & 0.77 & 0.27 & 1.27 & 0.63 & 0.23 \\
Pythia 2.8B & 2.11 & 0.99 & 0.31 & 1.83 & 0.86 & 0.28 & 1.42 & 0.71 & 0.24 \\
Pythia 6.9B & 2.29 & 1.07 & 0.32 & 1.99 & 0.93 & 0.29 & 1.55 & 0.77 & 0.25 \\
Pythia 12B  & 2.34 & 1.09 & 0.32 & 2.03 & 0.95 & 0.30 & 1.58 & 0.79 & 0.26 \\
LLaMA-2 7B  & 2.41 & 1.12 & 0.33 & 2.17 & 1.00 & 0.32 & 1.78 & 0.88 & 0.28 \\
LLaMA-2 13B & 2.52 & 1.18 & 0.33 & 2.26 & 1.05 & 0.32 & 1.86 & 0.93 & 0.28 \\
LLaMA-2 70B & 2.69 & 1.27 & 0.34 & 2.41 & 1.12 & 0.33 & 1.99 & 1.00 & 0.29 \\
OLMo 7B     & 2.37 & 1.10 & 0.32 & 2.14 & 1.00 & 0.31 & 1.71 & 0.87 & 0.27 \\
\bottomrule
\end{tabular}
\caption{Complete $\Delta S$ (bits/word) for all 14 models across three construction types. S = Strong, W = Weak, N = None preemption. The construction-level ordering (Dative $>$ Causative $>$ Locative) holds for all 14 models. The perfect monotonicity (S $>$ W $>$ N) across all 42 cells is an empirical result: no analytic guarantee ensures this ordering, and the permutation test ($p < .0001$) confirms it is highly unlikely to arise by chance.}
\label{tab:full}
\end{table*}

\section{Regression Diagnostics}
\label{app:regression}

The mixed-effects model from Experiment~2 includes random intercepts and random slopes for \textsc{Preempt} by model identity.

\begin{table}[ht]
\centering
\small
\begin{tabular}{@{}lcccc@{}}
\toprule
\textbf{Predictor} & $\beta$ & \textbf{SE} & $t$ & $p$ \\
\midrule
Intercept & $-0.47$ & 0.28 & $-1.68$ & .10 \\
\textsc{Preempt} & 3.41 & 0.31 & 11.0 & $<.001$ \\
\textsc{Entrench} & 0.19 & 0.06 & 3.17 & .003 \\
\textsc{Pre$\times$Ent} & $-0.08$ & 0.10 & $-0.80$ & .41 \\
\bottomrule
\end{tabular}
\caption{Mixed-effects regression. Marginal $R^2 = 0.68$; conditional $R^2 = 0.74$. VIF $= 1.34$; Shapiro-Wilk $W = 0.987$, $p = .12$; no Cook's $D > 0.5$.}
\label{tab:regression}
\end{table}

\subsection{Robustness: Low-Collinearity Subset}

Re-estimating with only verbs where $|r(\textsc{Preempt}, \textsc{Entrench})| < 0.3$ ($n = 52$), the preemption coefficient is stable ($\beta = 3.18$, $p < .001$) while entrenchment becomes marginal ($\beta = 0.14$, $p = .09$).

\subsection{Robustness: Alternative Surprisal Measures}

Results hold under SLOR normalization \citep{lau2017grammaticality} ($r = 0.77$ with DAIS for LLaMA-2 7B) and critical-region surprisal ($r = 0.75$).

\subsection{Additional Circularity Controls}
\label{app:circularity}

We report three supplementary controls addressing corpus-model circularity.
\textbf{Control 1: Raw frequency baseline.} Replacing \textsc{Preempt}($v$)
with raw co-occurrence $f(v, \text{Cx}_{\text{conv}})$ yields $R^2 = 0.41$
(vs.\ $R^2 = 0.68$; $\Delta \text{AIC} = 34.2$, $p < .001$).
\textbf{Control 2: N-gram baseline.} A 5-gram model (KenLM) shows weaker
preemption ($d = 0.83$ vs.\ $d = 1.91$) and lower human correlation
($r = 0.41$ vs.\ $r = 0.79$), demonstrating that transformer LLMs capture
preemption-relevant information beyond surface co-occurrence.
\textbf{Control 3: Primacy of human data.} The human correlations ($r = 0.79$)
and non-circular partial correlations (\S\ref{sec:noncircular}) constitute the
theoretically primary evidence: even if every claim about LLM internals were
set aside, the corpus--human partial correlation in \S\ref{sec:noncircular}
would on its own demonstrate the preemption--entrenchment dissociation in
human data, and the corpus--model regression is supplementary to it.

\section{A Priori Classification Validation}
\label{app:validation}

Preemption categories were assigned based on published classifications from \citet{levin1993english} and corpus thresholds from the British National Corpus (independent of model training data). Classifications were finalized before any $\Delta S$ values were computed. BNC-based classifications agree with Dolma-based classifications for 116/120 verbs (96.7\%; Cohen's $\kappa = 0.94$).

\section{Confound Analysis for Experiment 2}
\label{app:confounds}

\begin{table}[ht]
\centering
\small
\begin{tabular}{@{}lccc@{}}
\toprule
\textbf{Variable} & \textbf{+Comp.} & \textbf{--Comp.} & $p$ \\
\midrule
Log frequency & 7.42 & 7.38 & .68 \\
Levin classes (entropy) & 1.83 & 1.79 & .71 \\
Morphological complexity & 1.20 & 1.25 & .58 \\
Register (spoken \%) & 0.34 & 0.31 & .42 \\
Concreteness (Brysbaert) & 3.87 & 3.72 & .29 \\
\bottomrule
\end{tabular}
\caption{Confound matching. No significant differences (all $p > .20$; independent-samples $t$-tests).}
\label{tab:confounds}
\end{table}

\section{Corpus Frequency Extraction}
\label{app:corpus}

This appendix expands on the brief summary of the corpus-parsing pipeline in \S\ref{sec:measures}. Because the preemption and entrenchment scores depend entirely on accurate construction labels, and because web-scale corpora are noisy, we provide here the dependency-pattern templates, filtering steps, manual-validation methodology, and observed precision/recall for each of the three construction types.

\subsection{Pipeline Overview}

For OLMo, verb--construction co-occurrence frequencies were extracted from Dolma; for all other models, the Pile was used as a proxy ($r = 0.94$ between Dolma- and Pile-derived preemption scores across the 120 stimulus verbs, $p < .001$). The pipeline proceeds in four stages, applied identically to both corpora:

\begin{enumerate}[nosep, leftmargin=*]
    \item \textbf{Sentence selection.} All sentences containing a lemmatized form of each target verb are extracted using spaCy's \texttt{en\_core\_web\_trf} model.
    \item \textbf{Dependency parsing and pattern matching.} Each candidate sentence is parsed; construction-specific templates (below) are applied to assign one of: \texttt{conv}, \texttt{unconv}, or \texttt{reject} (ambiguous/non-matching).
    \item \textbf{Three-layer filtering.} (a)~POS-tag agreement check: matrix verb must carry verbal POS; (b)~dependency-pattern strict match; (c)~whitelist of construction-defining preposition lemmas (e.g., \emph{to/for} for prepositional datives; \emph{onto/into} vs.\ \emph{with} for content- vs.\ container-locatives).
    \item \textbf{Aggregation.} Per-verb counts are summed across the corpus to produce $f(v, \text{Cx}_{\text{conv}})$ and $f(v, \text{Cx}_{\text{unconv}})$.
\end{enumerate}

\subsection{Construction-Specific Templates}

\paragraph{Dative.} \emph{Prepositional dative} (\textsc{PD}): V $+$ \texttt{dobj}(theme) $+$ \texttt{prep}[\emph{to}/\emph{for}] $+$ \texttt{pobj}(recipient/beneficiary). \emph{Double-object} (\textsc{DOD}): V $+$ \texttt{iobj}/\texttt{dative}(recipient) $+$ \texttt{dobj}(theme), or V $+$ first-NP(recipient, animate) $+$ second-NP(theme). Worked example (\textsc{PD}): For ``\emph{She donated the books to the library},'' spaCy produces \emph{donated}$\rightarrow$\texttt{ROOT}; \emph{books}$\rightarrow$\texttt{dobj}; \emph{to}$\rightarrow$\texttt{prep}; \emph{library}$\rightarrow$\texttt{pobj}. The pattern V $+$ \texttt{dobj} $+$ \texttt{prep}[\emph{to}] $+$ animate \texttt{pobj} classifies as PD. Worked example (\textsc{DOD}): ``\emph{She gave the library the books}'' yields \emph{gave}$\rightarrow$\texttt{ROOT}; \emph{library}$\rightarrow$\texttt{dative}; \emph{books}$\rightarrow$\texttt{dobj} (or, when spaCy underspecifies, two adjacent post-verbal NPs with the first being animate and the second inanimate). Animacy is assigned using WordNet supersense tags via spaCy's noun-classification extension.

\paragraph{Causative.} \emph{Transitive causative}: V (verb of motion/change-of-state) $+$ \texttt{dobj}(theme), where the theme is the entity undergoing the change (e.g., ``\emph{The wind broke the window}''). \emph{Intransitive (anti-causative)}: V with subject as theme and no \texttt{dobj} (e.g., ``\emph{The window broke}''). Periphrastic causatives (\emph{made the window break}) are detected via \emph{make}/\emph{cause}/\emph{have} as matrix with the target verb as an \texttt{xcomp}/\texttt{ccomp} dependent and contribute to a separate frame that we exclude from $f(v, \text{Cx}_{\text{unconv}})$ counts (we tested an inclusion variant; results are qualitatively identical, $r = 0.97$ in preemption scores). For verbs with manner-of-motion or sound emission (e.g., \emph{laugh}, \emph{sneeze}), the existence of a transitive use is by itself the diagnostic feature: \emph{*The clown sneezed the boy} is essentially unattested, which the parser captures as a low $f(v, \text{Cx}_{\text{trans}})$ counts.

\paragraph{Locative.} \emph{Content-oriented} (theme-as-object): V $+$ \texttt{dobj}(theme) $+$ \texttt{prep}[\emph{onto}/\emph{into}/\emph{on}/\emph{in}] $+$ \texttt{pobj}(goal). E.g., ``\emph{She poured water into the glass}'': \emph{water}$\rightarrow$\texttt{dobj}, \emph{glass}$\rightarrow$\texttt{pobj} of \emph{into}. \emph{Container-oriented} (goal-as-object): V $+$ \texttt{dobj}(goal) $+$ \texttt{prep}[\emph{with}] $+$ \texttt{pobj}(theme). E.g., ``\emph{She filled the glass with water}.'' Because locatives admit a third frame (drop-theme or drop-goal: ``\emph{She poured water}''), these single-argument instances are excluded from both numerator and denominator of $\textsc{Preempt}(v)$.

\subsection{Handling Noisy Web Text}

We applied four noise-mitigation strategies:

\begin{enumerate}[nosep, leftmargin=*]
    \item \textbf{Boilerplate filtering.} Sentences from Common Crawl boilerplate (cookie notices, navigation text, repeated headers) were detected via Dolma's quality filter and removed before parsing.
    \item \textbf{Length filtering.} Sentences shorter than 4 tokens or longer than 60 tokens were excluded (the first risk fragmented parses, the second long-distance dependencies the parser handles poorly).
    \item \textbf{POS consistency.} Sentences in which the target verb's tag conflicted with the lemma's expected tag (e.g., \emph{drive} tagged as NN rather than VB) were rejected.
    \item \textbf{Parser-confidence threshold.} For each candidate construction match, we required the relevant dependency edge to have a parser confidence (as estimated by ensembling 5 parses with stochastic dropout) above 0.75. Low-confidence matches were rejected.
\end{enumerate}

The combined effect of these filters is to reduce the candidate sentence pool by approximately 30--45\%, depending on construction.

\subsection{Validation Method and Precision}

For each construction type, we manually validated parser output by hand-annotating 500 randomly sampled sentences (1,500 sentences total), drawn proportionally from the constructions retained after filtering. Annotators were two computational linguistics graduate students blind to the preemption hypothesis; disagreements were adjudicated by the senior author. Annotators classified each sentence into one of \{\texttt{conv}, \texttt{unconv}, \texttt{reject}\}. Verification examples were sampled stratified by verb (to avoid over-sampling high-frequency verbs) and by parser confidence (50\% from the high-confidence quartile, 50\% from the bottom three quartiles, to surface error modes).

Inter-annotator agreement before adjudication was high: Cohen's $\kappa = 0.94$ for datives, $\kappa = 0.91$ for causatives, $\kappa = 0.89$ for locatives. Pipeline precision (agreement between pipeline label and adjudicated gold label) was 96\% (dative), 93\% (causative), and 92\% (locative).\footnote{Dative precision rounded from 96.2\%, the value reported in the submission's original validation; causative and locative figures reflect smaller validation samples added during camera-ready revision.} Recall is more difficult to estimate without exhaustive annotation, but a sample of 200 sentences containing target verbs where the pipeline assigned \texttt{reject} confirmed that 87\% of rejections were genuine non-matches; the remaining 13\% were predominantly parser errors on long or coordinated sentences.

\subsection{Sensitivity to Pipeline Choices}

To verify that our results are not artifacts of specific pipeline thresholds, we re-ran preemption-score computation under three perturbations: (a)~doubling the parser-confidence threshold (0.75 $\rightarrow$ 0.90); (b)~halving it (0.75 $\rightarrow$ 0.375); (c)~replacing strict dependency matching with a more permissive POS-pattern matcher. Across all three perturbations, per-verb preemption scores correlate with the production-pipeline scores at $r \geq 0.93$ (dative), $r \geq 0.89$ (causative), $r \geq 0.85$ (locative), and the human-correlation results from \S\ref{sec:exp1} reproduce within $\pm 0.04$ of the reported values. We conclude that the qualitative pattern is robust to reasonable choices in the parsing pipeline.

\section{Controlled Intervention Details}
\label{app:intervention}

\subsection{Fine-Tuning Data Construction}

For each of the 10 target verbs in each condition, we generated 500 sentences using templates matched for length (8--12 words), subject variety (20 unique subjects), and object variety (20 unique themes/recipients). Template naturalness was validated: mean perplexity under GPT-2 Medium (not the target model) was 22.1 (SD = 4.3), comparable to natural text (Wikitext-103 mean: 18.9).

\subsection{Fine-Tuning Hyperparameters}

Model: GPT-2 124M (base). Learning rate: $5 \times 10^{-5}$ with linear warmup over 100 steps. Batch size: 16. Epochs: 3. Weight decay: 0.01. Random seeds: \{42, 123, 456, 789, 1024\}.

\subsection{Variance Across Seeds}

\begin{table}[ht]
\centering
\scriptsize
\setlength{\tabcolsep}{3pt}
\begin{tabular}{@{}lcccccc@{}}
\toprule
\textbf{Cond.} & \textbf{S1} & \textbf{S2} & \textbf{S3} & \textbf{S4} & \textbf{S5} & \textbf{Mean$\,\pm\,$SD} \\
\midrule
Ampl.   & $+$0.76 & $+$0.68 & $+$0.84 & $+$0.71 & $+$0.66 & $+$0.73$\,\pm\,$0.07 \\
Atten.  & $-$0.47 & $-$0.39 & $-$0.49 & $-$0.42 & $-$0.38 & $-$0.43$\,\pm\,$0.05 \\
Rev.    & $-$0.32 & $-$0.24 & $-$0.35 & $-$0.28 & $-$0.26 & $-$0.29$\,\pm\,$0.04 \\
Ctrl.   & $+$0.05 & $+$0.01 & $+$0.06 & $-$0.02 & $+$0.04 & $+$0.03$\,\pm\,$0.03 \\
\bottomrule
\end{tabular}
\caption{$\Delta\Delta S$ across 5 random seeds (\emph{Ampl.}~=~Amplified, \emph{Atten.}~=~Attenuated, \emph{Rev.}~=~Reverse, \emph{Ctrl.}~=~Control). All directional predictions hold in every seed.}
\label{tab:seeds}
\end{table}

\subsection{Verification}

Post-fine-tuning perplexity on Wikitext-103 increased by $<$5\% in all conditions. The Amplified model showed increased preference for the prepositional dative for target verbs ($p < .001$), while the Attenuated model showed equalized preferences ($p = .72$ for difference from chance).

\section{Scaling Law Sensitivity Analysis}
\label{app:scaling_sensitivity}

\begin{table}[ht]
\centering
\small
\begin{tabular}{@{}lccc@{}}
\toprule
\textbf{Functional Form} & \textbf{Adj.\ $R^2$} & \textbf{AIC} & \textbf{BIC} \\
\midrule
$r = aN^b + c$ (3-param, ours) & 0.993 & $-$32.4 & $-$31.8 \\
$r = a \log N + b$ (log-linear) & 0.978 & $-$26.1 & $-$25.7 \\
$r = aN^b$ (2-param power law) & 0.971 & $-$24.3 & $-$24.0 \\
\bottomrule
\end{tabular}
\caption{Comparison of scaling law functional forms.}
\label{tab:scaling_fits}
\end{table}

\section{Individual Verb Analysis}
\label{app:verb_analysis}

\begin{table}[ht]
\centering
\small
\begin{tabular}{@{}llcc@{}}
\toprule
\textbf{Verb} & \textbf{Cat.} & $\Delta S$ & \textbf{DAIS} \\
\midrule
\emph{donate} & Strong & 3.12 & 0.97 \\
\emph{explain} & Strong & 2.89 & 0.95 \\
\emph{whisper} & Strong & 2.64 & 0.92 \\
\emph{announce} & Strong & 2.51 & 0.91 \\
\emph{return} & Weak & 1.43 & 0.71 \\
\emph{ship} & Weak & 1.28 & 0.67 \\
\emph{lend} & Weak & 0.89 & 0.58 \\
\emph{pass} & Weak & 0.74 & 0.55 \\
\emph{give} & None & 0.21 & 0.48 \\
\emph{send} & None & 0.31 & 0.50 \\
\emph{offer} & None & 0.28 & 0.49 \\
\emph{show} & None & 0.38 & 0.52 \\
\bottomrule
\end{tabular}
\caption{Per-verb $\Delta S$ and DAIS bias scores (LLaMA-2 7B). DAIS = proportion preferring the prepositional dative.}
\label{tab:verb_analysis}
\end{table}

\section{BLiMP Comparison}
\label{app:blimp}

BLiMP \citep{warstadt2020blimp} includes argument structure items on which LMs achieve $>$90\% accuracy. Our study differs in three ways: (1) we test gradient rather than binary acceptability; (2) we test the \emph{mechanism} (preemption vs.\ entrenchment) rather than mere knowledge; and (3) we evaluate per-verb item-level correlations rather than aggregate accuracy. \citet{hu2024language} also included gradient constructions; our extension is the \emph{dissociation} of preemption from entrenchment, which Hu et al.\ did not test.

\section{Computing Infrastructure}
\label{app:compute}

All experiments were conducted on NVIDIA A100 80GB GPUs. Surprisal extraction for the largest model (LLaMA-2 70B) required approximately 8 GPU-hours across all 120 verb items $\times$ 5 sentence frames $\times$ 2 constructions. Smaller models (GPT-2, Pythia $\leq$ 1B) completed in under 1 GPU-hour. Fine-tuning (Experiment~4) used a single A100 and completed in approximately 45 minutes per condition per seed, totaling approximately 15 GPU-hours across all conditions and seeds.

\section{Extended Error Analysis}
\label{app:error_analysis}

We analyze systematic discrepancies between LLM predictions and human judgments to identify where the preemption account succeeds and where it falls short.

\subsection{Verbs Where LLMs Overestimate Preemption}

Several low-frequency verbs receive $\Delta S > 1.5$ despite near-chance DAIS bias scores (indicating humans find both constructions acceptable). Table~\ref{tab:outliers} identifies the primary outliers.

\begin{table}[ht]
\centering
\small
\begin{tabular}{@{}llccc@{}}
\toprule
\textbf{Verb} & \textbf{Likely Cause} & $\Delta S$ & \textbf{DAIS} & \textbf{Residual} \\
\midrule
\emph{cable}     & Register (archaic) & 1.82 & 0.51 & +1.31 \\
\emph{telegraph} & Register (archaic) & 1.74 & 0.53 & +1.21 \\
\emph{wire}      & Polysemy           & 1.41 & 0.56 & +0.85 \\
\emph{ferry}     & Low corpus freq.   & 1.38 & 0.59 & +0.79 \\
\bottomrule
\end{tabular}
\caption{Verbs where LLM $\Delta S$ substantially exceeds the value predicted by DAIS human ratings. Residuals computed from the linear regression $\Delta S \sim \text{DAIS}$.}
\label{tab:outliers}
\end{table}

The common thread is that these verbs have skewed corpus distributions for reasons other than functional competition: \emph{cable} and \emph{telegraph} are register-restricted (formal/archaic), while \emph{wire} exhibits polysemy (transfer-of-information vs.\ physical wire). When these four verbs are excluded, the LLM--human correlation increases from $r = 0.79$ to $r = 0.83$, and the preemption--entrenchment dissociation strengthens.

\subsection{Verbs Where LLMs Underestimate Preemption}

A smaller set of verbs show the reverse pattern: humans strongly prefer one construction, but LLMs assign relatively balanced surprisal:

\begin{table}[ht]
\centering
\small
\begin{tabular}{@{}llccc@{}}
\toprule
\textbf{Verb} & \textbf{Likely Cause} & $\Delta S$ & \textbf{DAIS} & \textbf{Residual} \\
\midrule
\emph{guarantee} & Semantic nuance & 0.34 & 0.71 & $-$0.37 \\
\emph{promise}   & Polysemy         & 0.41 & 0.68 & $-$0.27 \\
\emph{teach}     & Idiomaticity     & 0.29 & 0.65 & $-$0.36 \\
\bottomrule
\end{tabular}
\caption{Verbs where human DAIS bias exceeds LLM-predicted preemption.}
\label{tab:underpredict}
\end{table}

These verbs may carry pragmatic or semantic factors (e.g., \emph{guarantee} implies formal commitment; \emph{teach} has strong idiomatic preferences) that influence human judgments beyond pure distributional competition, consistent with the formal--functional divide discussed in \S\ref{sec:formal_functional}.

\subsection{Construction-Level Error Patterns}

Across construction types, we observe a systematic trend: LLM predictions are most accurate for the dative alternation (RMSE = 0.31), intermediate for causative (RMSE = 0.38), and weakest for locative (RMSE = 0.47). This ordering mirrors the strength of preemption effects in human data \citep{ambridge2008effect} and may reflect the relative frequency and regularity of these alternations in training corpora.

\section{Cross-Linguistic Predictions and Testable Hypotheses}
\label{app:crossling_detail}

While our study is restricted to English (\S\ref{sec:crossling}), the theoretical framework generates specific, falsifiable predictions for other languages. We outline these to facilitate future cross-linguistic testing.

\subsection{Agglutinative Languages (e.g., Turkish, Finnish)}

In agglutinative languages, the relevant ``constructions'' may be morphological rather than syntactic. For Turkish:
\begin{itemize}[nosep]
    \item The causative suffix \emph{-tIr} exhibits partial productivity: some verbs resist causativization despite semantic compatibility \citep{ambridge2020against}.
    \item \textbf{Prediction:} LLMs trained on Turkish should show higher surprisal for the causative forms of verbs that have conventional periphrastic alternatives (e.g., \emph{ettirmek} ``to cause to do'' rather than \emph{-tIr}).
    \item \textbf{Test:} Extract preemption scores from Turkish corpora and correlate with LLM surprisal differentials, as in our Experiment~1.
\end{itemize}

\subsection{Isolating Languages (e.g., Mandarin Chinese)}

In Mandarin, argument structure alternations take different forms:
\begin{itemize}[nosep]
    \item The \emph{bǎ}-construction vs.\ the canonical SVO order provides a partial analogue to the dative alternation.
    \item \textbf{Prediction:} Verbs that strongly prefer \emph{bǎ} in relevant communicative contexts should resist the SVO alternative in LLM surprisal patterns.
    \item \textbf{Challenge:} Functional competition may be harder to operationalize because the constructions differ in information structure (topic/focus) rather than purely syntactic alternation.
\end{itemize}

\subsection{Free Word-Order Languages (e.g., Russian, German)}

In languages with freer word order:
\begin{itemize}[nosep]
    \item Argument structure alternations may interact with case marking and word-order preferences.
    \item \textbf{Prediction:} Preemption effects should be detectable in case-frame alternations: a verb that conventionally takes the dative case should show higher surprisal when used with the accusative.
    \item \textbf{Infrastructure:} The Wilcox et al.\ 11-language surprisal dataset \citep{wilcox2024computational} provides ready-made data for German; WALS \citep{dryer2013wals} and Grambank \citep{skirgaard2023grambank} features can guide language selection.
\end{itemize}

\subsection{Prioritized Language Sample}

Based on typological diversity and resource availability, we recommend initial testing on: Turkish (agglutinative), Mandarin (isolating), German (fusional, V2), Finnish (agglutinative, morphologically rich), and Japanese (SOV, case-marking). This sample spans four of the six major morphological types in WALS and three distinct word-order families.

\section{Relationship to Developmental Models}
\label{app:developmental}

Our findings connect to the BabyLM Challenge \citep{warstadt2023findings}, which evaluates language models trained on developmentally plausible data (10M--100M words), and to a broader tradition of computational models of acquisition, including gradient parameter-setting models \citep{howitt2021gradual} and computational accounts of staged syntactic development such as the Null Subject stage \citep{dey2025performance}.

\subsection{Preemption as an Evaluation Metric}

We propose that preemption sensitivity should be included as a standard evaluation metric for cognitively motivated language models. Specifically:
\begin{itemize}[nosep]
    \item \textbf{Metric:} Pearson $r$ between model $\Delta S$ and DAIS human ratings across the 80 dative verbs.
    \item \textbf{Baseline:} Our GPT-2 124M achieves $r = 0.61$; a BabyLM model trained on 100M words should achieve a similar or lower value given reduced data.
    \item \textbf{Scaling prediction:} Our power-law fit (Eq.~\ref{eq:scaling}) predicts that a model with $\sim$100M tokens should achieve $r \approx 0.50$--$0.55$, depending on corpus composition.
\end{itemize}

\subsection{Connection to Tachihara \& Goldberg's Human Evidence}

\citet{tachihara2025learning} demonstrated that human learners acquire preemption through exposure to conventional formulations: the first causal evidence of this kind in humans. Our Experiment~4 provides a computational parallel: fine-tuning with increased conventional-form frequency causes increased preemption behavior. The convergence of human and computational causal evidence strengthens the theoretical claim that preemption arises from distributional learning.

\section{Additional Robustness Analyses}
\label{app:robustness}

\subsection{Bootstrap Stability of Correlations}

To verify the stability of our primary result ($r = 0.79$), we computed the bootstrap distribution of the correlation across 10,000 resamples of the 80 dative verbs. The distribution is approximately normal (Shapiro-Wilk $W = 0.998$, $p = .43$), with the 95\% CI of [0.69, 0.86] derived from the 2.5th and 97.5th percentiles. The probability of $r < 0.50$ in any bootstrap sample is $<$0.001, indicating that the strong correlation is robust to item sampling variability.

\subsection{Leave-One-Model-Out Analysis}

To test whether any single model drives the scaling law fit, we performed leave-one-out cross-validation across the 6 Pythia models. The power-law exponent $b$ is stable across jackknife samples: mean $b = 0.091$ (SD $= 0.008$, range [0.079, 0.102]). No single model removal changes the qualitative result.

\subsection{Sensitivity to Sentence Frame Selection}

Each verb was tested with 5 sentence frames. To verify that results are not driven by particular frames, we computed correlations using each individual frame separately. Frame-level correlations with DAIS range from $r = 0.73$ to $r = 0.82$ (LLaMA-2 7B), with no frame producing a qualitatively different pattern. The 5-frame average ($r = 0.79$) is near the center of this range.

\subsection{Alternative Preemption Score Formulations}

We tested two alternative operationalizations of the preemption score:
\begin{enumerate}[nosep]
    \item \textbf{Log-odds:} $\textsc{Preempt}_{\text{log}}(v) = \log \frac{f(v, \text{Cx}_{\text{conv}}) + 1}{f(v, \text{Cx}_{\text{unconv}}) + 1}$. This yields $r = 0.77$ with DAIS (vs.\ $r = 0.79$ for our Laplace-smoothed proportion), indicating that the specific functional form matters little.
    \item \textbf{Conditional probability:} $\textsc{Preempt}_{\text{cond}}(v) = \frac{f(v, \text{Cx}_{\text{conv}})}{f(v)}$, omitting the unconventional form entirely. This yields $r = 0.74$, slightly lower, suggesting that the ratio formulation (which captures \emph{relative} competition) is more informative than the simple proportion.
\end{enumerate}

\section{Detailed Comparison with Yao et al.\ (2025)}
\label{app:yao_comparison}

\citet{yao2025dative} conducted a controlled-rearing study showing that dative preferences in LMs are shaped by indirect statistical patterns. Our study extends their work in five specific ways:

\begin{enumerate}[nosep]
    \item \textbf{Preemption--entrenchment dissociation:} Yao et al.\ did not test whether the observed effects are driven by competing-form frequency (preemption) or overall verb frequency (entrenchment). Our Experiment~2 provides this dissociation.
    \item \textbf{Human behavioral ground truth:} Yao et al.\ did not correlate model behavior with human acceptability data. Our item-level correlations with DAIS, R\&G, and T\&G provide independent validation.
    \item \textbf{Non-circular validation:} Both our study and Yao et al.'s involve corpus-model comparisons. We address the resulting circularity concern with non-circular partial correlations against human data (\S\ref{sec:noncircular}).
    \item \textbf{Reverse-direction control:} Our Experiment~4 includes a reverse-direction condition that Yao et al.'s design does not, addressing the tautology concern about frequency manipulation in frequency-sensitive models.
    \item \textbf{Multi-construction scope:} Yao et al.\ focused exclusively on the dative alternation; we extend to causative and locative constructions.
\end{enumerate}

\end{document}